\documentclass[journal]{IEEEtran}
\ifCLASSINFOpdf
\else
\fi
\usepackage{amsmath}
\usepackage{graphicx}
\usepackage{multirow}
\usepackage{array}
\usepackage{booktabs}

\usepackage{url}

\usepackage{hyperref}
\usepackage{etoolbox}
\makeatletter
\patchcmd{\@makecaption}
  {\scshape}
  {}
  {}
  {}
\makeatother
\begin{document}
%
\title{Vehicle Lane Change Prediction based on Knowledge Graph Embeddings and Bayesian Inference}
%
%
%

\author{M. Manzour$^{1}$, A. Ballardini$^{1}$, R. Izquierdo$^{1}$, and M. A. Sotelo$^{1}$\\
$^{1}$Department of Computer Engineering, University of Alcal\'a, Madrid, Spain\\
$[$ahmed.manzour, augusto.ballardini, ruben.izquierdo, miguel.sotelo$]$@uah.es}
\maketitle

\begin{abstract}
Prediction of vehicle lane change maneuvers has gained a lot of momentum in the last few years. Some recent works focus on predicting a vehicle's intention by predicting its trajectory first. This is not enough, as it ignores the context of the scene and the state of the surrounding vehicles (as they might be risky to the target vehicle). Other works assessed the risk made by the surrounding vehicles only by considering their existence around the target vehicle, or by considering the distance and relative velocities between them and the target vehicle as two separate numerical features. In this work, we propose a solution that leverages Knowledge Graphs (KGs) to anticipate lane changes based on linguistic contextual information in a way that goes well beyond the capabilities of current perception systems. Our solution takes the Time To Collision (TTC) with surrounding vehicles as input to assess the risk on the target vehicle. Moreover, our KG is trained on the HighD dataset using the \textit{TransE} model to obtain the Knowledge Graph Embeddings (KGE). Then, we apply Bayesian inference on top of the KG using the embeddings learned during training. Finally, the model can predict lane changes two seconds ahead with 97.95\% f1-score, which surpassed the state of the art, and three seconds before changing lanes with 93.60\% f1-score.
\end{abstract}

\begin{IEEEkeywords}
Lane Change Prediction, Knowledge Graph Embeddings, Bayesian Inference, Bayesian Reasoning
\end{IEEEkeywords}

%
\IEEEpeerreviewmaketitle

\section{Introduction}
%
%
%
%
\IEEEPARstart{A}{ccidents} occur every day in our daily lives, and the number of deaths due to vehicle crashes is increasing every year. Based on statistics published in $2023$ by the National Highway Traffic Safety Administration (NHTSA), the number of deaths in motor vehicle traffic crashes in the United States of America (USA) in $2021$ is $42000$. Which is a $10\%$ increase in the number of deaths compared to $2020$, and a $17.3\%$ increase compared to $2019$ \cite{stewart2023overview}. Lane-changing maneuvers are one of the causes of vehicle crashes, as a report indicated that $33\%$ of all road crashes happen due to the existence of a vehicle that changes its lane. Also, the NHTSA indicated the fact that $94\%$ of vehicle crashes are the driver’s fault \cite{singh2015critical}. That’s why the government put some constraints on the road and the driver, like wearing seat belts and being committed to the road speed limit. Also, researchers started to focus on implementing different models to predict the vehicle lane-changing intention to reduce the number of accidents/crashes on the road.

Most of the recently proposed models are based on target vehicle trajectory data and certain relative measurements with the surrounding vehicles (e.g. relative distances and velocities). Furthermore, these models are based on numerical input values, which makes them act like a black box. This makes reasoning and interpreting the model outputs difficult. also, it is challenging to explain the model and its outputs to others who may not be familiar with the underlying algorithms. So, this work focuses on addressing the following points:

\begin{enumerate}
    \item Lane change prediction is carried out based on contextual information, not merely using kinematic information learned from previous experiences. This goes beyond the capabilities of current perception systems and allows to generalize and make predictions agnostic to the physical aspect of the road environment.
    
    \item Predictions are based on knowledge graphs and, consequently, they are interpretable and explainable, contributing to developing trustworthy systems. 

    \item Bayesian inference is carried out as a downstream task on the grounds of the learned embeddings, allowing the implementation of a fully inductive reasoning system based on KGEs.
\end{enumerate} 

\begin{figure}[t]
\centering
\includegraphics[width=\columnwidth]{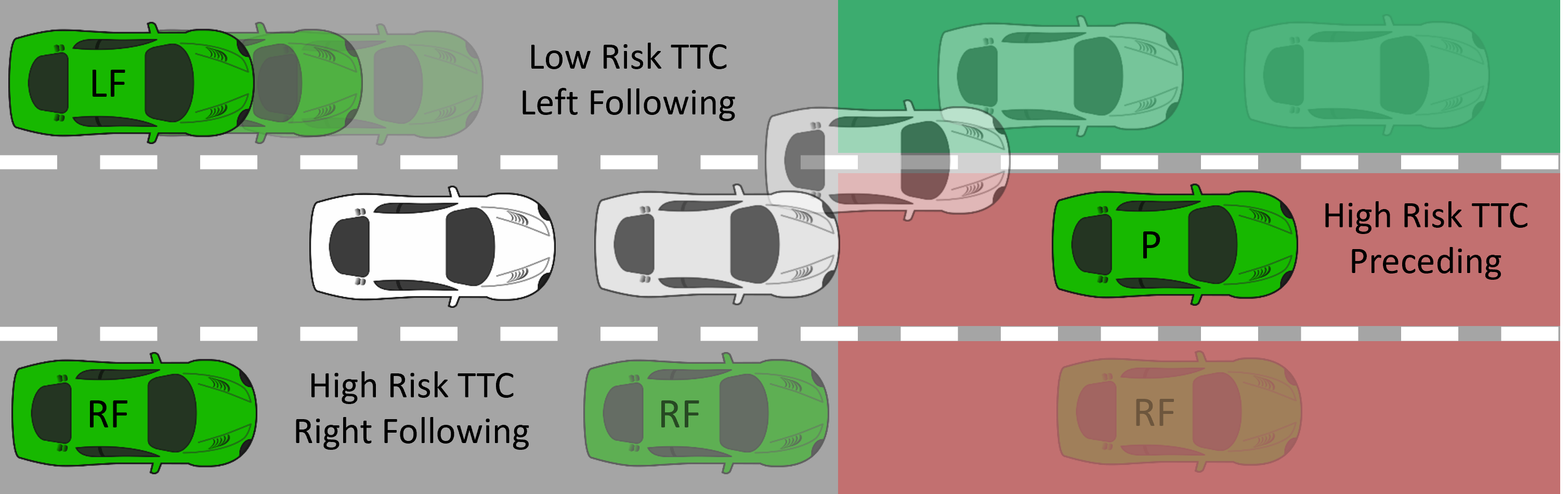}
\caption{Target (white) vehicle will make left lane-changing maneuver based on the risk assessment of the surrounding (green) vehicles.}
\label{fig:introduction_lane_changing_example}
\end{figure}

The inputs that are fed to the model can describe the risk situation around the target vehicle in a linguistic manner so everyone can understand and reason why the target vehicle took a certain maneuver. For example, \autoref{fig:introduction_lane_changing_example} shows a scenario of a white target vehicle that will make a left lane change because there is high-risk TTC with the preceding vehicle (P) and high-risk TTC with the right following vehicle (RF) as well. Still, there is low-risk TTC with the left following vehicle (LF). So, the target vehicle will avoid lane keeping or right lane changing in order to avoid collisions/risks with the preceding vehicle or the right following vehicle, respectively. Instead, the target vehicle will execute the left lane changing maneuver.

The rest of this article is organized as follows. \autoref{sec:sota} presents the state of the art. \autoref{sec:preliminary} contains a brief introduction to the Highway Drone (HighD) dataset and Knowledge Graphs (KGs). Then, our proposed methodology will be discussed in detail in \autoref{sec:methodology}. In \autoref{sec:results}, results will be presented. Finally, \autoref{sec:concliusions} concludes the work and provides some recommended future work.

\section{State of the Art}\label{sec:sota}
Recently, different works have focused on predicting vehicle lane changes using different inputs (including target vehicle position, speed, acceleration, and surrounding vehicles' states) and methods (like rule-based algorithms and data-driven models). Rule-based algorithms define certain rules at which the vehicle makes a lane change, such as the gap acceptance model, which assumes that the vehicle will make a lane change when it achieves the least possible distance between it and the front/rear vehicles in the same lane as in \cite{yang1996microscopic}. Data-driven models are based on training a model on certain inputs to obtain a certain complex equation that maps these inputs to a certain output. The models can be some traditional machine learning models like Support Vector Machine (SVM) and logistic regression, or deep learning models like Artificial Neural Networks (ANN) and Recurrent Neural Networks (RNN) with all its variations.

In $2017$, authors in \cite{woo2017lane} used SVM and artificial potential field models to detect lane changes for vehicles based on predicting their trajectories using the Next-Generation Simulation (NGSIM) dataset. They also used the predicted trajectories to consider the possibility of a crash with an adjacent vehicle to reduce false alarms. The inputs are the distance from the centerline, the lateral velocity, and the potential feature. The output is whether the vehicle will change lanes or keep lanes. The work concluded that most zigzag driving cases are originally lane-changing maneuvers. However, these maneuvers are canceled due to the presence of a vehicle in the adjacent lane.

In $2018$, \cite{su2018learning} utilized a Long Short-Term Memory (LSTM) model to predict vehicle lane changes by considering the vehicle's past trajectory and neighbors' states. The models take different inputs extracted from the NGSIM dataset. The inputs were the vehicle's lateral and longitudinal global positions with respect to the lane, the vehicle's acceleration, the existence of right/left lane vehicles, and the longitudinal distance between the target vehicle and (front/rear)(left, center, right) surrounding vehicles.

Work \cite{benterki2019prediction} in $2019$ utilized two machine learning models to predict lane changes of surrounding vehicles on highways. The inputs were extracted from the NGSIM dataset. The inputs were longitudinal/lateral velocities, longitudinal/lateral accelerations, distance to left/right lane markings, yaw angle, and yaw rate related to the road. These inputs were trained and tested on SVM and ANN models.

In $2019$, authors in \cite{izquierdo2019experimental} predicted lane-changing intentions of surrounding vehicles using two different methodologies and by only using visual information provided by the PREVENTION dataset. The first method was Motion History Image - Convolutional Neural Network (MHI-CNN), where temporal and visual information was obtained from the MHI, and then fed to the CNN model. The second model was the GoogleNet-LSTM model, in which a feature vector was obtained from a GoogleNet CNN model and then fed to the LSTM model to learn temporal patterns. The used inputs were the RGB image, center (X, Y), and the bounding box's dimensions (W, H). The results showed that the GoogleNet-LSTM model outperformed the MHI-CNN model.

In $2020$, \cite{laimona2020implementation} trained LSTM and RNN models on the PREVENTION dataset to predict surrounding vehicles' lane-changing intentions by tracking the vehicles' positions (centroid of the bounding box). Sequences of $10$, $20$, $30$, $40$, and $50$ frames of (X, Y) coordinates of the target vehicle were considered for comparison. It was concluded that RNN models performed better on short sequence lengths and the LSTM model outperformed RNN at long sequences.

The work implemented in \cite{xue2022integrated} in $2022$ utilized eXtreme Gradient Boosting (XGBoost) and LSTM to predict the vehicle lane change decision and trajectory prediction, respectively in scenarios in the HighD dataset. The models were based on the traffic flow (traffic density) level, the type of vehicle, and the relative trajectory between the target vehicle and surrounding vehicles. At first, the traffic flow and vehicle type models were separately implemented. The traffic flow $y_t$ model took the longitudinal velocity $v_{lon}$ and acceleration $a_{lon}$ of the target vehicle, Headway, and the relative velocity between the target vehicle and the (left/front/right preceding, and left/right following vehicles). The vehicle type $y_v$ model took all the inputs stated previously concatenated with the angle between the target vehicle trajectory and road vertical line $\varphi$. Then, the lane change decision prediction was achieved utilizing the XGBoost model. The model took the following inputs: $v_{lon}$, $a_{lon}$, $\varphi$, lateral velocity $v_{lat}$ and acceleration $a_{lat}$ of the target vehicle, Headway and the relative velocity between the target vehicle and the mentioned five surrounding vehicles, $y_t$, and $y_v$. Finally, vehicle trajectory prediction occurred based on historical trajectories and the predicted lane-changing decision.

In $2023$, \cite{gao2023dual} built a dual transformer which contained two transformer models. One was for lane change prediction, while the other was for trajectory prediction. The first model used the target vehicle's historical lateral trajectory information and the surrounding vehicles' states, including the longitudinal distance and velocity between the target vehicle and (the left/front/right preceding vehicle and the left/right following vehicles). The intention prediction output obtained from the first model was fused with the target vehicle's historical lateral trajectory information and fed to the second model to establish the connection between the intentions and the trajectories. The dual transformer was trained and validated on the HighD and NGSIM datasets. Finally, this research assumed that the ego vehicle sensors can obtain all the information regarding the position and speed of the target vehicle and all the surrounding vehicles. 

After closely examining the previous literature, the following research gaps can be identified, which are covered throughout this work.
\begin{enumerate}
    \item Lane change prediction based on contextual linguistic information, not only using trajectory numerical information learned from previous experiences. 
    
    \item Predictions are interpretable and explainable as they are based on KGs. 

    \item Enabling a fully inductive reasoning system using Bayesian inference built on top of the KGEs.
\end{enumerate} 

\section{Preliminary}\label{sec:preliminary}
This section briefly introduces the dataset used in this work (HighD), and the foundation model (KG and KGE) used to develop the lane change prediction system.

\subsection{HighD Dataset}\label{subsec:highd}
The HighD dataset \cite{highDdataset} is a German dataset that records naturalistic top-view scenes on German highway roads using a camera integrated into a quadcopter. The dataset is recorded in six locations and includes $60$ tracks ($\approx$ $15$ minutes each) containing more than $110,500$ vehicles. For each vehicle in the dataset, much information is provided, including but not limited to vehicle position, speed, acceleration, TTC with the preceding (front) vehicle, whether a left/center/right preceding, left/right side, or left/center/right following (rear) vehicle exists, and the vehicle's current lane.  The dataset can be reached through the following link: \url{https://levelxdata.com/highd-dataset/}.
\subsection{Knowledge Graph and Knowledge Graph Embedding}\label{subsec:kgs}
A graph is a type of database that represents complex data in a structured way, which is human-interpretable and can be easily observed and analyzed. It stores information in the form of entities (nodes) and their relationships (edges). Each relation connects two nodes and acts as the relationship between them. The relation can be directed (x isFatherOf y) or undirected (x isFriendWith y, y isFriendWith x).\cite{zheng2020dgl}

A knowledge graph is a directed heterogeneous (nodes can have different types) multigraph (each pair of nodes can be connected with more than one relation). In the context of KGs, a triple $<s,r,o>$ consists of subject and object entities connected by a relation. For example, $<$\textit{vehicle}, \textit{INTENTION\textunderscore IS}, \textit{leftLaneChanging}$>$. The subject is \textit{vehicle}, the relation is \textit{INTENTION\textunderscore IS}, and the object is \textit{leftLaneChanging}.\cite{zheng2020dgl}

KGE is a supervised machine learning task that learns to represent (embed) the knowledge graph entities and relations into a low-dimensional vector space while preserving semantic meaning. There are several KGE models, such as DistMult, TransE, RotatE, ComplEx, and HolE. Each model's unique scoring function measures the distance between two entities using the relation between them. The purpose of the scoring function is to make entities connected by a relation close to each other in the vector space, while entities that do not belong to this relation should be far apart. \cite{zheng2020dgl, ampligraph}

\section{Methodology}\label{sec:methodology}
\subsection{Architecture Overview} \label{subsec:arquitecture}
\begin{figure*}[!h]
\centering
\includegraphics[width=\linewidth]{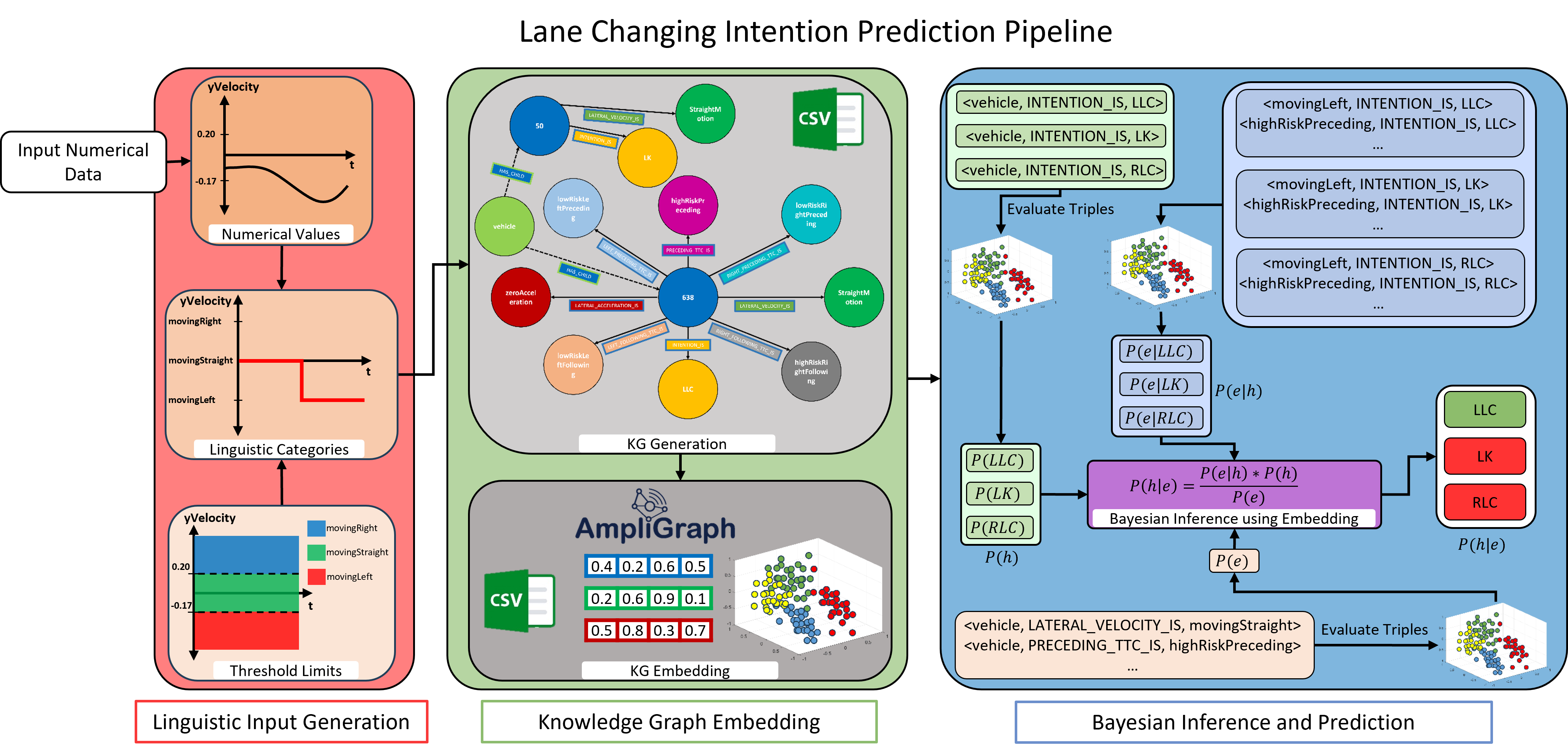}
\caption{The pipeline for anticipating lane changes consists of three phases: Linguistic Input Generation, Knowledge Graph Embedding, and Bayesian Inference and Prediction.}
\label{fig:intention_prediction_pipeline}
\end{figure*}

\autoref{fig:intention_prediction_pipeline} shows the pipeline of our proposed methodology. The pipeline consists of three phases. Phase one is the linguistic input generation phase, in which the numerical input variables are converted to linguistic input categories using some threshold limits. Phase two is the KGE phase in which the KG will be generated as triples in a CSV file and embedded (trained) using the Ampligraph library \cite{ampligraph}. The third phase is the Bayesian inference and prediction phase. This phase is responsible for calculating the probability that a vehicle will make a \textit{LLC}, \textit{LK}, and \textit{RLC} given the linguistic inputs generated from phase one. And the highest probability is the model prediction. The calculation of these probabilities is based on the formation of triples and the evaluation of these triples using the embedding obtained from phase two and the Bayesian reasoning in phase three. The KG ontology and input structure will be stated in the next section. Then, each of the three phases is discussed in detail.

\subsection{Knowledge Graph Ontology and Input Definition}
The KG ontology is a formal (general) representation of the entities and their relationships in the KG. In KGs, ontologies are important because they act as a schema for constructing the KG so that they can ensure consistency and explainability of the KG. Also, It is worth mentioning that we apply a number of reifications on the given inputs to get reified triples.

\begin{figure}[h]
\centering
\includegraphics[width=\columnwidth]{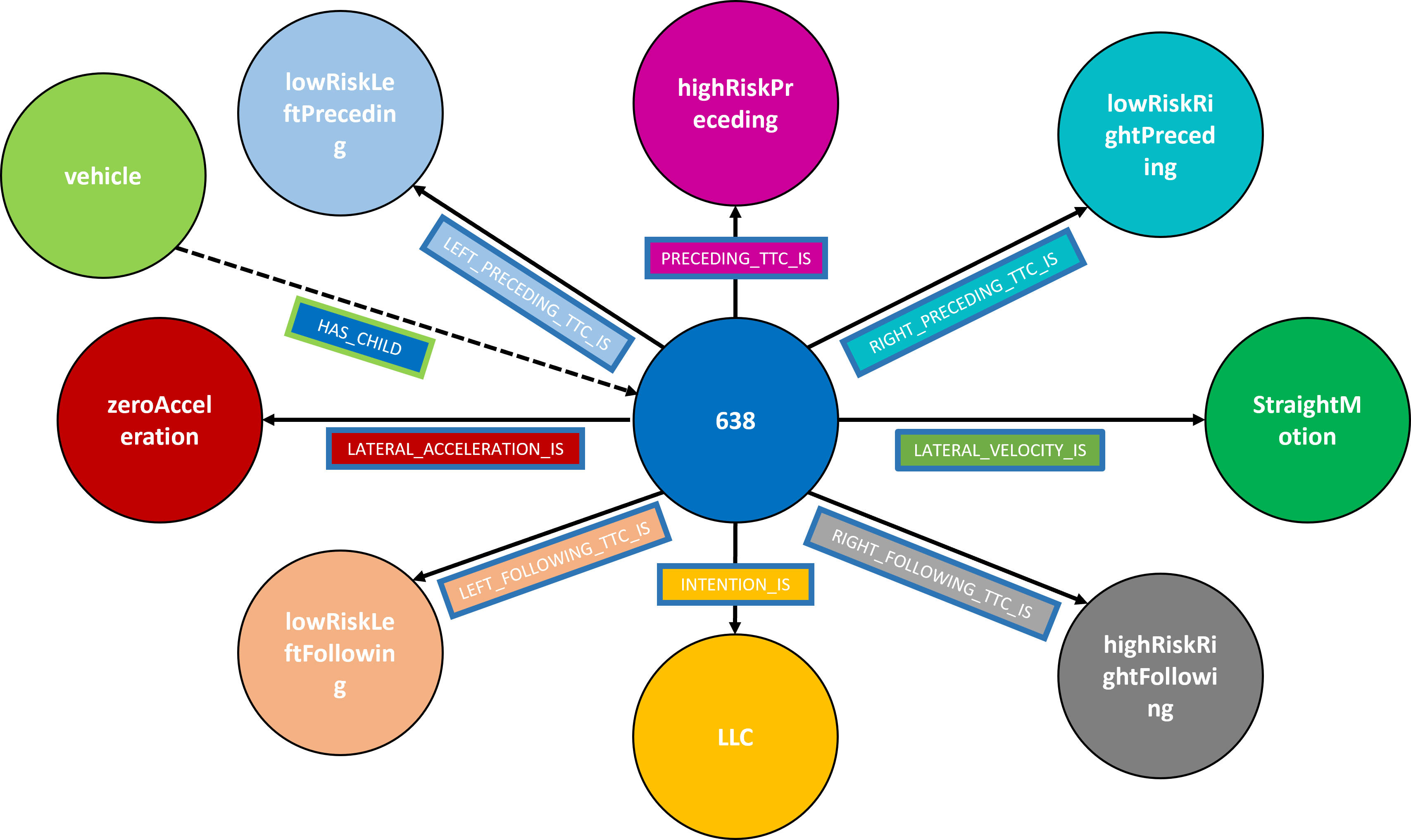}
\caption{One KG instance where every entity (class) is assigned to its unique instance.}
\label{fig:KG_instance}
\end{figure}

For example, if the target vehicle has a preceding vehicle, and the TTC with this preceding vehicle is at high risk, then, the reified triple is $<$\textit{vehicle}, \textit{PRECEDING\textunderscore TTC\textunderscore IS}, \textit{highRiskPreceding}$>$ where \textit{vehicle} in that triple points to the target vehicle. \autoref{tab:KG_classes} shows the ontology for our lane change prediction KG model. The table shows all the input/output classes in the first column, a description in the second column, a set of possible reified linguistic instances (categories) for each class in the third column, a description of that instance (if needed) in the fourth column, and the name of the relation pointing to that class after reification in the fifth column. \autoref{fig:KG_instance} shows a KG instance based on the formed ontology. For example, it can be observed that the generic entity \textit{vehicle} has a child with ID \textit{638}, and this child has \textit{latVelocity} class assigned to \textit{movingStraight} instance. Also, this child has \textit{highRiskPreceding} TTC. \textit{638} intention is \textit{LLC}.

\begin{figure}[h]
\centering
\includegraphics[width=\columnwidth]{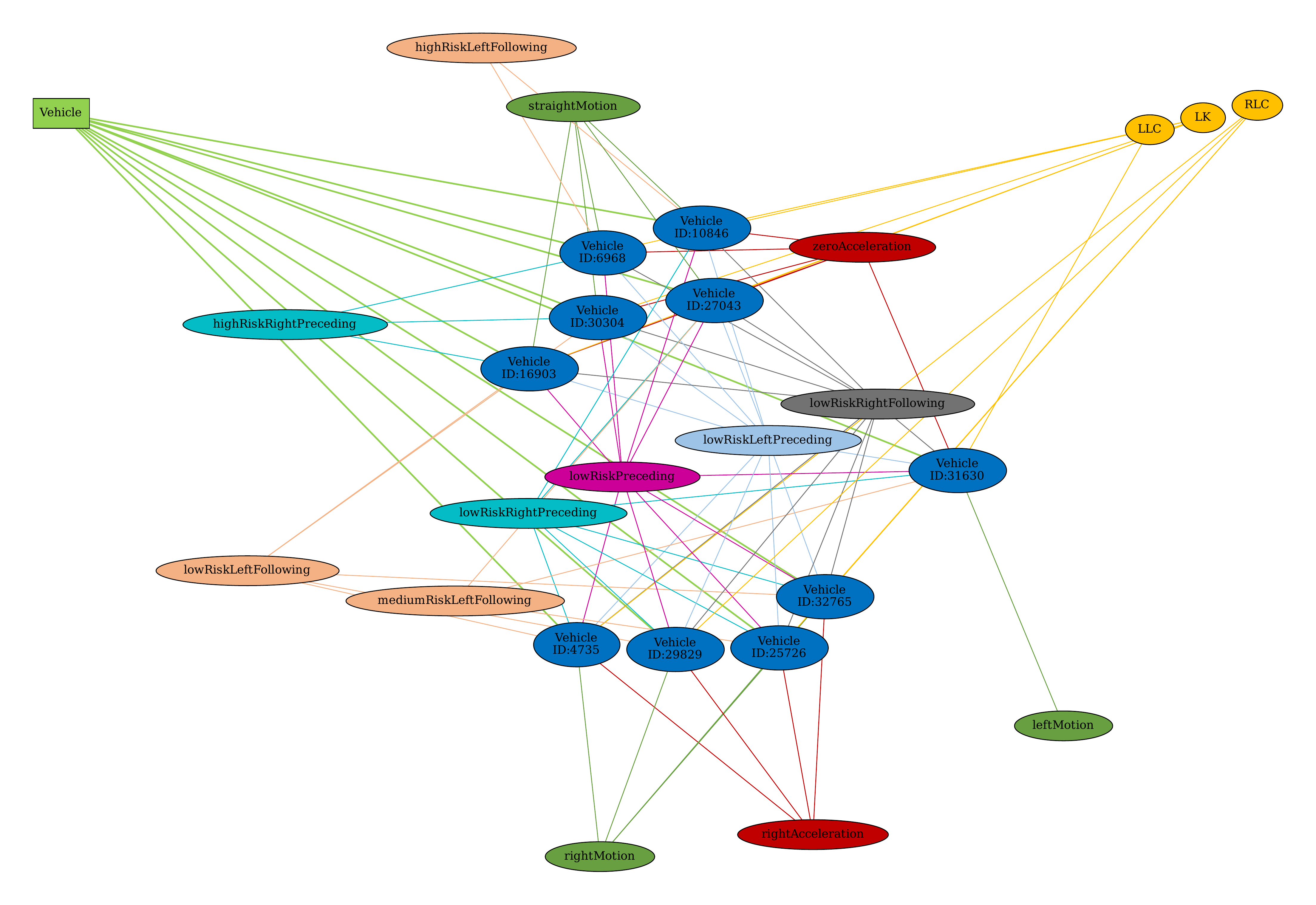}
\caption{KG with 10 instances where the vehicle generic entity is connected to 10 child vehicles.}
\label{fig:v3_all_vehicles_10_neato}
\end{figure}

\begin{figure}[h]
\centering
\includegraphics[width=\columnwidth]{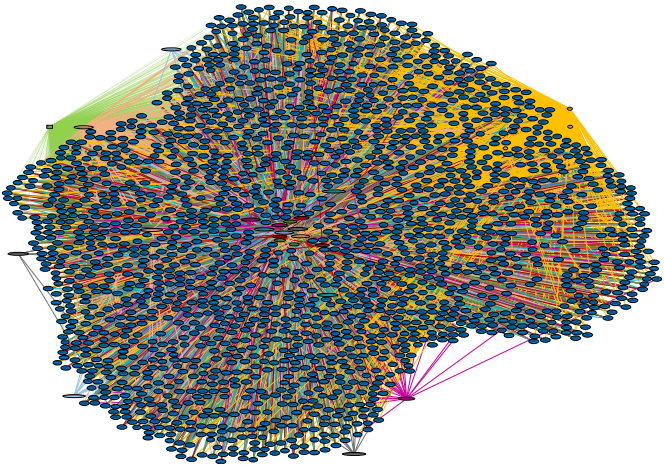}
\caption{Large KG with 2000 instances from a total of 39304 instances.}
\label{fig:v3_all_vehicles_2000_neato}
\end{figure}

The same applies to all the entities in the graph for this child and any other child. Note that the entity \textit{vehicle} is connected to all other children, which forms a big KG with many instances connected to each other through that generic entity as shown in \autoref{fig:v3_all_vehicles_10_neato} and \autoref{fig:v3_all_vehicles_2000_neato}. Such that \autoref{fig:v3_all_vehicles_10_neato} contains KGs for only 10 vehicles, and \autoref{fig:v3_all_vehicles_2000_neato} contains KG instances for 2000 vehicles. It is worth mentioning that the total number of instances used for training and validation is 39304, which is hard to fit into one figure.
\renewcommand*{\arraystretch}{1.2}
\begin{table*}[h]
\centering
\caption{Ontology table which includes the definition of all entities (classes), their instances, and possible relations that can be connected to them.}
\label{tab:KG_classes}
\begin{tabular}{|c|>{\centering\arraybackslash}m{3cm}|c|c|c|}

\hline
Class & Class Description & Instance & Instance Description & Possible Relation \\ \hline\hline

&   & LLC & Left Lane Changing   &\\ 
intention & Lane changing intention & LK & Lane Keeping & INTENTION\textunderscore IS \\ 
& of the vehicle & RLC & Right Lane Changing  &\\ \hline\hline

& & movingLeft & -- &\\ 
latVelocity & Vehicle lateral velocity & movingStraight & -- & LATERAL\textunderscore VELOCITY\textunderscore IS\\ 
& & movingRight & -- &\\ \hline\hline

& & leftAcceleration & -- &\\ 
latAcceleration & Vehicle lateral accelera- & zeroAcceleration & No lateral acceleration & LATERAL\textunderscore ACCELERATION\textunderscore IS\\ 
& tion & rightAcceletion & -- &\\ \hline\hline

& & highRiskPreceding & -- &\\ 
ttcPreceding & TTC with the preceding  & mediumRiskPreceding & -- & PRECEDING\textunderscore TTC\textunderscore IS \\ 
& (front) vehicle & lowRiskPreceding & -- &\\ \hline\hline

& & highRiskLeftPreceding & -- &\\ 
ttcLeftPreceding & TTC with the left & mediumRiskLeftPreceding & -- & LEFT\textunderscore PRECEDING\textunderscore TTC\textunderscore IS\\ 
& preceding (front) vehicle & lowRiskLeftPreceding & -- &\\ \hline\hline

& & highRiskRightPreceding & -- &\\ 
ttcRightPreceding & TTC with the right & mediumRiskRightPreceding & -- & RIGHT\textunderscore PRECEDING\textunderscore TTC\textunderscore IS \\ 
& preceding (front) vehicle & lowRiskRightPreceding & -- &\\ \hline\hline

& & highRiskLeftFollowing & -- &\\ 
ttcLeftFollowing & TTC with the left & mediumRiskLeftFollowing & -- & LEFT\textunderscore FOLLOWING\textunderscore TTC\textunderscore IS \\ 
& following (rear) vehicle & lowRiskLeftFollowing & --  &\\ \hline\hline

& & highRiskRightFollowing & -- &\\ 
ttcRightFollowing & TTC with the right & mediumRiskRightFollowing & -- & RIGHT\textunderscore FOLLOWING\textunderscore TTC\textunderscore IS \\ 
& following (rear) vehicle & lowRiskRightFollowing & -- &\\ \hline\hline

vehicleID & Child vehicle ID which changes every frame & vehicle ID number (e.g. '638') & -- & HAS\textunderscore CHILD\\ \hline\hline

vehicle & Generic entity pointing to every child vehicle & -- & -- & Any\\ \hline

\end{tabular}
\end{table*}

\subsection{Linguistic Input Generation Phase} \label{subsec:conversion}
The used inputs to the model are vehicle lateral (velocity and acceleration), TTC to (preceding, left preceding, right preceding, left following, right following) vehicles. The first three inputs are directly provided by the dataset. The TTC inputs for the preceding left and right vehicles are extracted based on the following equation. That $d$ is the distance between the two vehicles. The sub-letter $p$ refers to the left or right preceding vehicle. Given that the distance $d$ and both velocities are provided by the dataset.
\begin{equation}
    TTC = \frac{d_{p}}{v_{target}-v_{p}}
\end{equation}

The TTC inputs for the left and right following vehicles are extracted based on the following equation. The sub-letter $f$ refers to the left or right following vehicle.
\begin{equation}
    TTC = \frac{d_{f}}{v_{f}-v_{target}}
\end{equation}

After extracting all the needed features. All the numerical data is converted to linguistic 'string' categories. 

\begin{figure}[!h]
\centering
\includegraphics[width=\columnwidth]{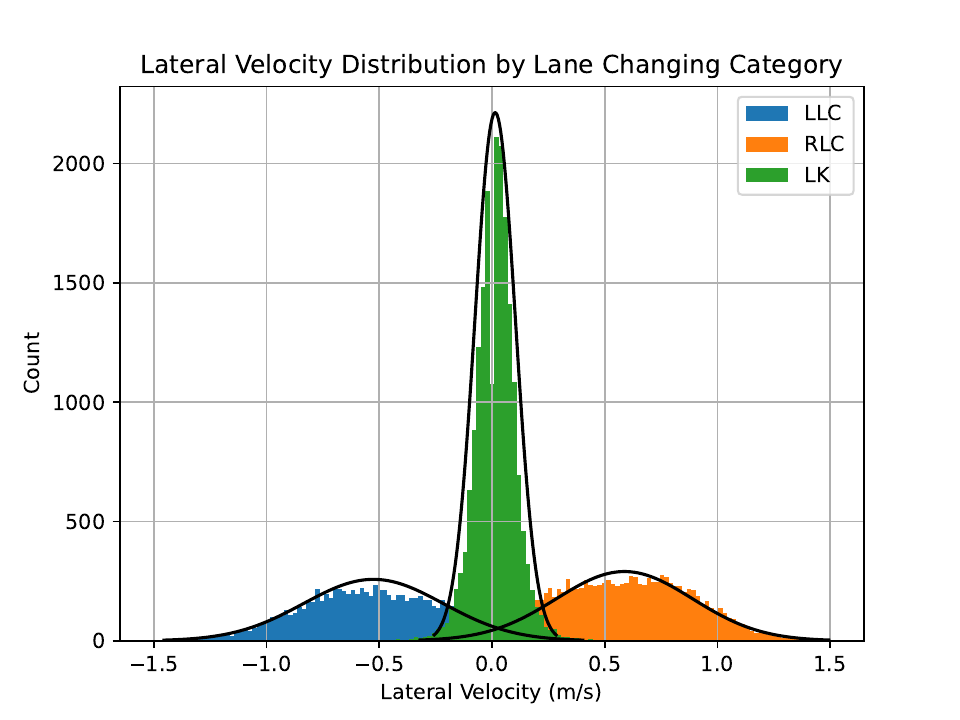}
\caption{Histogram shows the relation between lateral velocity numerical values and the lane-changing labels.}
\label{fig:lateral_velocity_histogram}
\end{figure}

The conversion from numerical value to a linguistic category is done by taking each feature and converting it to a string based on some threshold limits obtained from statistics and literature. For example, the \textit{latVelocity} feature numerical values (shown in  \autoref{fig:lateral_velocity_histogram}) can be separated into three linguistic categories $\{$\textit{movingLeft}, \textit{movingStraight}, \textit{movingRight}$\}$. 

The challenge here is to choose the numerical threshold limit values that separate the three linguistic categories accurately. These values are obtained based on the normal distribution of the data shown in \autoref{fig:lateral_velocity_histogram}. Based on the standard deviation $\sigma$ and mean $\mu$ for the data of each lane changing category, the $\mu \pm 2\sigma$ values represent the threshold limit of each linguistic variable. So, in this case, the \textit{movingStraight} linguistic variable has a threshold limit $[\mu-2\sigma, \mu+2\sigma]$, \textit{movingLeft} $(-\infty, \mu-2\sigma)$, and \textit{movingRight} $(\mu+2\sigma, \infty)$.

\begin{figure}[!h]
\centering
\includegraphics[width=\columnwidth]{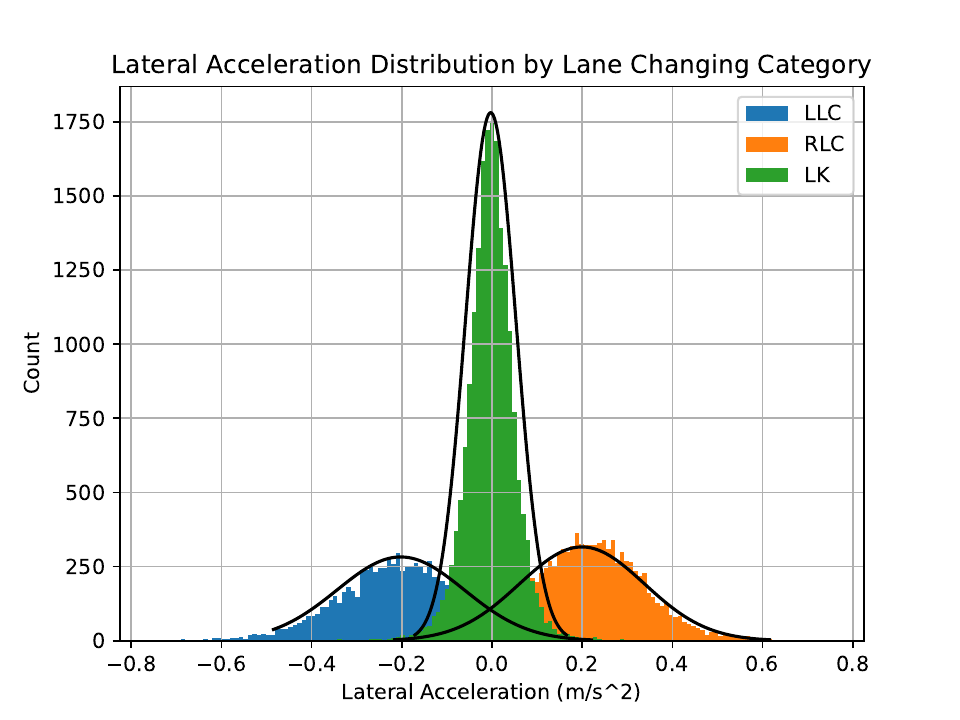}
\caption{Histogram shows the relation between lateral acceleration numerical values and the lane-changing labels.}
\label{fig:lateral_acceleration_histogram}
\end{figure}

For the lateral acceleration, \autoref{fig:lateral_acceleration_histogram} shows the histogram of the lateral acceleration numerical values. Based on the data distribution, the same category separation criteria (as in lateral velocity) is applied to obtain the lateral acceleration linguistic categories.

\begin{figure}[!h]
\centering
\includegraphics[width=\columnwidth]{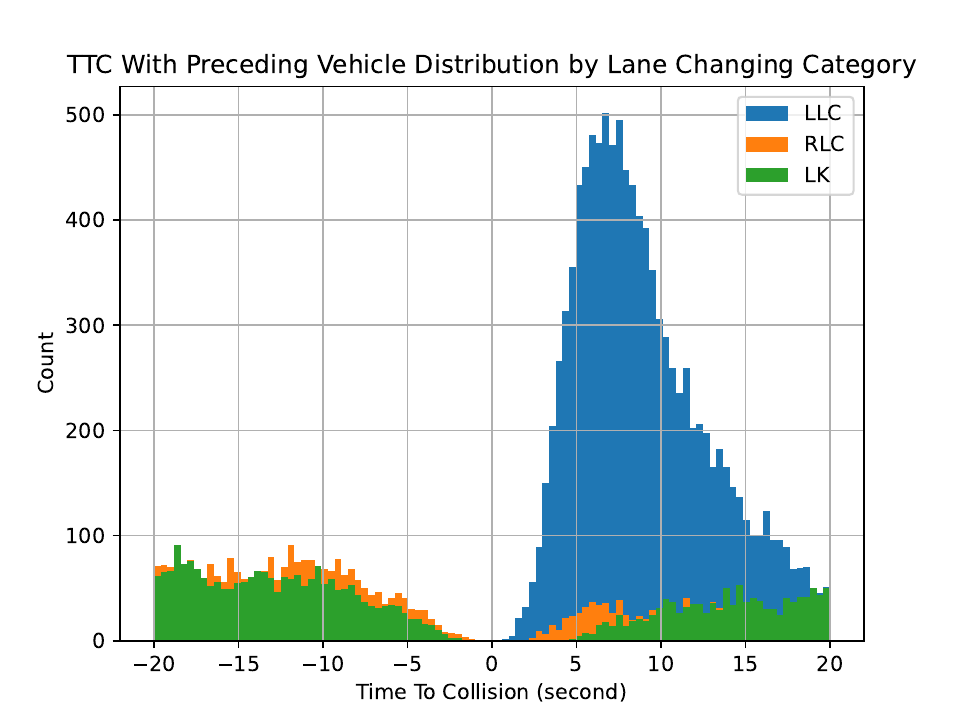}
\caption{Histogram shows the relation between numerical values of TTC with the preceding vehicle and the lane-changing labels.}
\label{fig:ttc_preceding_histogram}
\end{figure}

Regarding the TTC variables \autoref{fig:ttc_preceding_histogram} shows the data distribution for TTC with the preceding vehicle. It can be observed that at very low TTC values, there are no vehicles. That is because, at low positive TTC values near zero, the target vehicle is near and faster than the preceding vehicle, and there is a high risk of collision while approaching the preceding vehicle. So, vehicles prefer to avoid that risk. For negative values near zero, this means that the preceding vehicle is very near but faster than the target vehicle, and any breaking action taken by the preceding vehicle will cause a low positive TTC which might lead to a collision too. So, most vehicles prefer to be safe and avoid being in a region corresponding to these values. For the other TTC values in the plot outside the interval around zero, some vehicles keep lanes or change lanes to the right. But most vehicles change lanes to the left at TTC approximate range of $[4, 10]$ seconds. This interval is within the range that indicates risk on the target vehicle while following a preceding vehicle as stated by \cite{ramezani2020comparing} and considered in the study conducted by \cite{saffarzadeh2013general}. So, the vehicle changes lanes to the faster lane, which most probably is the left lane. Moreover, as shown in the figure, separating the TTC into categories is hard. Also, \cite{saffarzadeh2013general} indicated that drivers’ behavior is inconsistent in different situations; there is no definitive value for TTC threshold limits to enable classification between safe and unsafe maneuvering situations. Such that, \cite{ramezani2020comparing} stated that TTC values can be divided into three-time segments that show the correlation between the TTC values and the driver braking behavior which is used in risk assessment of the situation. TTC values between zero and four seconds indicate high correlation, TTC values between four and 16 seconds indicate low correlation, and TTC values larger than 16 indicate negligible correlation. \cite{ramezani2020comparing} used TTC values between 0.5 and 10 seconds to discriminate between safe and unsafe car-following situations during the study of improving the TTC formulation. So, in this work, the selected TTC thresholds for each surrounding vehicle location are based on the TTC with preceding vehicle values provided by the mentioned works. Therefore, TTC $\in [0,4]$ is \textit{high-risk}, TTC $\in (4,10)$ is \textit{medium-risk}, and any other positive or negative TTC value is \textit{low-risk}.

\subsection{Knowledge Graph Embedding Phase}
After the linguistic input generation phase, the KG is formed in a CSV file in the form of triples. Then, trained using the Ampligraph library.
\subsubsection{Knowledge Graph Generation}
As mentioned, the KG is generated in the form of triples in a CSV file. The file contains three columns and many rows. Each row represents an entity and its relationship with another entity. The structure of each row consists of three parts: subject (first column), predicate (second column), and object (third column). Returning to the example in \autoref{fig:KG_instance}, the triples CSV file will have the following structure. $<$\textit{vehicle}, \textit{HAS\textunderscore CHILD}, \textit{638}$>$, this is the first row. $<$\textit{638}, \textit{INTENTION\textunderscore IS}, \textit{LLC}$>$, this is the second row. Then, the formation of the triple will go by using the same procedure with the other nodes. Note that each vehicle will have a new ID in each frame even if it is the same vehicle. So, in the next frame, $<$\textit{vehicle}, \textit{HAS\textunderscore CHILD} \textit{639}$>$. \textit{638} and \textit{639} are IDs for the same vehicle, but during triples and KG generation, they are considered as IDs of two different vehicles. This graph can be extended to all the indicated seven inputs and to any number of vehicles by following the same structure.

\subsubsection{Knowledge Graph Embedding}
After producing the triples CSV file, Ampligraph 2.0.1 library \cite{ampligraph} is used for KGE. Training/validation and testing data are separated based on tracks to ensure that the vehicles' behavior is not overlapping, as vehicles from the same track can have similar behavior (e.g. track has a right exit at the end of the road). So, The first 48  (80\%) tracks in the dataset are used for training and validation. The other 12 (20\%) tracks are kept for testing. Different numbers of triples are used for validation (500, 1K, 2K, 4K, and 10K). However, they all had the same results during testing. So, only 2K triples are considered for validation. These triples are provided by using the \textit{train\textunderscore test\textunderscore split\textunderscore no\textunderscore unseen} function provided by the Ampligraph library. The numbers of triples are $351736$, $2000$, and $12222$ for training, validation, and testing, respectively. The distribution of the number of samples for each lane change category after tracks separation and triple generation is shown in \autoref{tab:data_distribution}.

 \begin{table}[ht]
\renewcommand{\arraystretch}{1.5}
\caption{Number of samples for each lane change category in training, validation, and testing sets.}
\begin{center}
\begin{tabular}{|l|c|c|c|}
\hline
     & LLC & LK & RLC \\
\hline
Training   & 8906  & 19109 & 11071\\
\hline
Validation  & 54  & 91 & 73\\
\hline
Testing    & 261  & 792 & 305\\
\hline
\end{tabular}
\label{tab:data_distribution}
\end{center}
\end{table}

Two scoring models were tested: \textit{TransE} and \textit{ComplEx}. Training parameters are fixed for fair comparison: embedding size $k=100$, five negative triples are generated for each positive triple where both the subject and the object of triples are corrupted, Adam optimizer with $learning\textunderscore rate=0.0005$, $SelfAdversarialLoss$, $batch\textunderscore size=10000$, $validation\textunderscore burn\textunderscore in=5$, $validation\textunderscore freq=5$, $validation\textunderscore batch\textunderscore size=100$. Finally, early stopping criteria is used to monitor $val\textunderscore mrr$ with the patience of $5$ validation epochs in order to stop training when there is no improvement in the $val\textunderscore mrr$.

\subsection{Bayesian Inference and Prediction Phase}

After training, the embeddings for every entity will become available. Our proposed solution is intended to allow for inductive reasoning. For that purpose, we implement some reifications in the graph (at the ontology level) and we carry out Bayesian inference on the learned embeddings. 

After obtaining the embeddings, we can compute the probabilities of the reified triples using the KGE evaluation method provided by the AmpliGraph library. These triples have the form $P(h,r,t)$, where h is the head (or subject entity), r is the relation, and t is the tail (or object entity). Based on that, $P(h \|e)$ (\textit{h} stands for hypothesis and \textit{e} stands for evidence) is computed using Bayes rule as follows\autoref{eq:bayesian}:

\begin{equation}
\begin{aligned}
    P(h \|e)=\frac{P(h)P(e\|h )}{P(e)}
\end{aligned}
    \label{eq:bayesian}
\end{equation}
where hypothesis \textit{h} is the event or entity that we want to predict (lane changing prediction), and evidence \textit{e} is the information that we have measured with onboard sensors for the current frame, which are the inputs given by the HighD dataset in that case. For example, hypothesis: vehicle lane changing intention is left, evidence: the vehicle TTC with the preceding vehicle is risky and the vehicle is accelerating laterally to the left.

Computation of $P(h)$ takes place by evaluating a single triple after reification as in this example: (vehicle lane changing intention is left) can be reified to $<$\textit{vehicle}, \textit{INTENTION\textunderscore IS}, \textit{LLC}$>$.

$P(e)$ is computed using the following equation:
\begin{equation}
    P(e) = P(e_1) \times  P(e_2) \times … \times P(e_n)
    \label{eq:evidence}
\end{equation}
where each $P(e_i)$ can also be computed by reification in the graph. For example, evidence 1 ($e_1$), which says that the vehicle TTC with the preceding vehicle is risky, can be reified as $<$\textit{vehicle}, \textit{PRECEDING\textunderscore TTC \textunderscore IS}, \textit{highRiskPreceding}$>$; evidence 2 ($e_2$) which says that vehicle is accelerating laterally to the left can be reified as $<$\textit{vehicle}, \textit{LATERAL\textunderscore ACCELERATION \textunderscore IS}, \textit{leftAcceleration}$>$. All pieces of evidence are reified following this philosophy.

Regarding the computation of $P(e\|h)$, It can be rewritten as the following: 

\begin{equation}
\begin{aligned}
     P(e\|h) = P(e_1\; and\;  e_2\; and\; …\; and\; e_n \|h)\\
     =P(e_1 | h) \times … \times P(e_n | h)\\
     =P(e_{1c}) \times … \times P(e_{nc})
\end{aligned}   
    \label{eq:evidence-hypothesis}
\end{equation}
 where $P(e_{ic})$ stands for the probability of $e_i$ given that the hypothesis \textit{h} is true. Also, conditioned pieces of evidence are reified. For the given example, $e_{1c}$ is: what is the probability of having a \textit{highRiskPreceding} vehicle, given that the hypothesis is that the target vehicle makes \textit{LLC}. This can be reified to the triple $<$\textit{highRiskPreceding}, \textit{INTENTION\textunderscore IS}, \textit{LLC}$>$. It means that we take it for granted that the object entity is a vehicle, and it is for sure changing its lane to the left lane. In such conditions, the probability that such a vehicle in such circumstances will have a risky preceding vehicle will be computed. Same applies to $e_{2c}$, $<$\textit{leftAcceleration}, \textit{INTENTION\textunderscore IS}, \textit{LLC}$>$. Then, $e_{1c}$ is multiplied by $e_{2c}$ to get $P(e\|h)$ as in \autoref{eq:evidence-hypothesis}. Finally, $P(h \|e)$ can be calculated using \autoref{eq:bayesian} given that all these individual probabilities are computable from the graph using the embeddings. 

 \begin{table*}[ht]
\renewcommand{\arraystretch}{1.5}
\caption{Comparison with other models using the f1-score (\%) metric.}
\begin{center}
\begin{tabular}{|l|c|c|c|c|c|c|c|c|c|}
\hline
Prediction time & 0.5 Seconds & 1 Second & 1.5 Seconds & 2 Seconds & 2.5 Seconds & 3 Seconds & 3.5 Seconds & 4 Seconds\\
\hline
$\cite{xue2022integrated}$   & 98.20  & 97.10  & 96.61 & 95.19 & $-$ & $-$ & $-$ & $-$\\
\hline
$\cite{gao2023dual}$   & \underline{\textbf{99.18}}  & \underline{\textbf{98.98}}  & 97.56 & 91.76 & $-$ & $-$ & $-$ & $-$\\
\hline
Ours     & 97.72  & 97.86  & \underline{\textbf{98.11}} & \underline{\textbf{97.95}} & 97.21 & 93.60 & 82.77 & 66.52\\
\hline
\end{tabular}
\label{tab:comparison}
\end{center}
\end{table*}

\section{Results}\label{sec:results}
This section presents the results achieved by our proposed model. The machine used to carry out this experiment is a Lenovo Legion laptop with Windows 11, i7-9750H CPU, 32GB of RAM, and NVIDIA GeForce RTX 2070 with Max-Q Design GPU. The KGE TransE model prediction time is $0.065$ seconds/prediction. The whole architecture prediction time is $0.455$ seconds/prediction.

After embedding the KG and utilizing Bayesian reasoning to get the predictions, the model is tested on the last 12 tracks in the HighD dataset. Testing started by comparing the f1-score of the \textit{TransE} model and the \textit{ComplEx} model three seconds before changing lanes. The \textit{TransE} model has $93.60\%$ f1-score, and the \textit{ComplEx} model scores $12\%$ f1-score. So, the \textit{TransE} model worked better and all the upcoming experiments and discussions will be based on using the \textit{TransE} model.

Testing takes place at different instances before crossing the lane line starting with $0.5$ seconds till four seconds with a step of $0.5$ seconds. \autoref{tab:results} shows the results at one, two, three, and four seconds. It can be observed that the model maintains the f1-score percentage over $90\%$ for the first three seconds. Then, the performance starts to drop till it reaches 6$6.52\%$ f1-score four seconds before changing lanes. After that, the model is compared in terms of f1-score with other literature that used the HighD dataset; the comparison is shown in \autoref{tab:comparison}. The table shows that \cite{xue2022integrated}, and \cite{gao2023dual} have an average of $1\%$ score margin with the proposed model at $0.5$ and $1$ seconds. Starting from $1.5$ seconds our model still maintains its performance with approximately the same f1-score and passes both models as their scores start to decrease. Our model kept the f1-score higher than $97\%$ for the first $2.5$ seconds, higher than $90\%$ for three seconds, and higher than $80\%$ for $3.5$ seconds before crossing the lane marking.

\begin{table}[ht]
\renewcommand{\arraystretch}{1.5}
\caption{Precision, recall, and f1-score metrics of the predictions obtained from our proposed model at different instants.}
\begin{center}
\begin{tabular}{rrrrr}
    1 Second          & \textbf{Precision (\%)} & \textbf{Recall (\%)} & \textbf{F1-score (\%)}\\
\hline
LK              & 98.33  & 96.96  & 97.64 \\
LLC             & 97.98  & 97.50  & 97.74 \\
RLC             & 97.00  & 99.42  & 98.19 \\
\hline
Macro avg       & 97.77  & 97.96  & 97.86 \\
\hline\hline
     2 Seconds           & \textbf{Precision (\%)} & \textbf{Recall (\%)} & \textbf{F1-score (\%)}\\
\hline
LK              & 98.86  & 96.96  & 97.95 \\
LLC             & 97.50  & 99.15  & 98.32 \\
RLC             & 96.52  & 98.66  & 97.58 \\
\hline
Macro avg       & 97.66  & 98.25  & 97.95 \\
\hline\hline
     3 Seconds           & \textbf{Precision (\%)} & \textbf{Recall (\%)} & \textbf{F1-score (\%)}\\
\hline
LK              & 92.53  & 96.96  & 94.70 \\
LLC             & 95.71  & 91.77  & 93.70 \\
RLC             & 95.46  & 89.50  & 92.38 \\
\hline
Macro avg       & 94.56  & 92.74  & 93.60 \\
\hline\hline
     4 Seconds           & \textbf{Precision (\%)} & \textbf{Recall (\%)} & \textbf{F1-score (\%)}\\
\hline
LK              & 69.63  & 96.96  & 81.05 \\
LLC             & 91.30  & 46.00  & 61.16 \\
RLC             & 88.75  & 42.39  & 57.37 \\
\hline
Macro avg       & 83.22  & 61.78  & 66.52 \\
\end{tabular}
\label{tab:results}
\end{center}
\end{table}


\begin{figure*}[h]
\centering
\includegraphics[width=\linewidth]{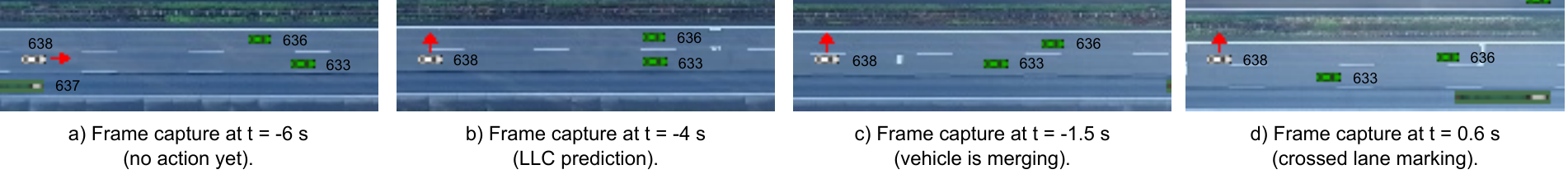}
\caption{Scene explanation through four different frame captures.}
\label{fig:frames_LLC_49_0638}
\end{figure*}


\begin{figure}[h]
\centering
\includegraphics[width=\columnwidth]{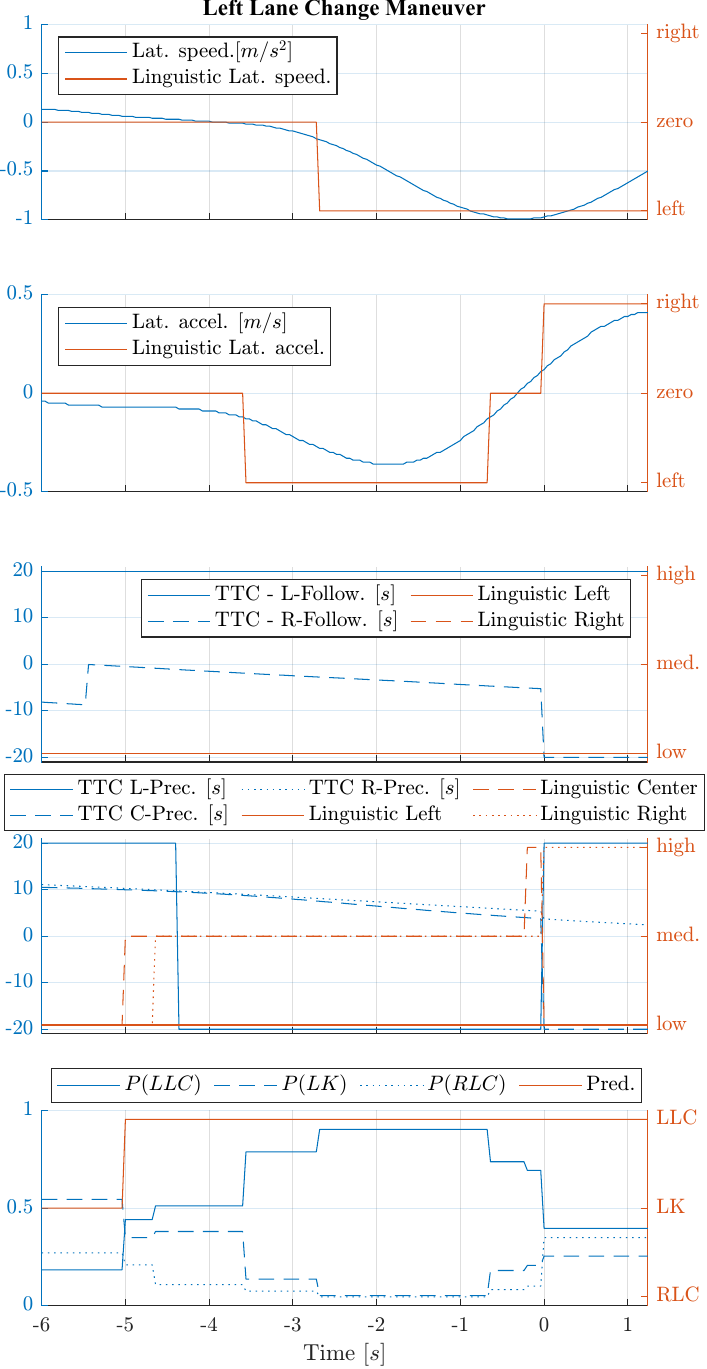}
\caption{Temporal sequence of numerical variables and linguistic categories.}
\label{fig:results_LLC_49_0638}
\end{figure}

\autoref{fig:frames_LLC_49_0638} and \autoref{fig:results_LLC_49_0638} shows left lane-changing scenario, where \autoref{fig:frames_LLC_49_0638} shows different captures at different instants for the white target vehicle and its surrounding green neighbors in that scenario. While \autoref{fig:results_LLC_49_0638} shows the numerical values and linguistic categories of the inputs that are fed to the KG model for the same scene. The upper sub-figure is for the lateral velocity, followed by the lateral acceleration, and then the TTC with left/right following vehicles. After that, the TTC with the left/center/right preceding vehicles. Finally, the last sub-figure is for the prediction probabilities throughout the scene. The focus here is to show that the model uses human/machine interpretable and explainable linguistic inputs to get a reasonable prediction.

The scene starts with \autoref{fig:frames_LLC_49_0638}a at $-6$ seconds before changing the lane. By observing the plots in \autoref{fig:results_LLC_49_0638}, the target vehicle has zero lateral velocity (\textit{movingStraight}), zero lateral acceleration, and low-risk TTC with the left/center/right preceding and left/right following vehicles. At this moment, human reasoning says that the vehicle is keeping its lane. Using Bayesian reasoning, the model is asked to compute the probability of \textit{LLC} given the generated linguistic inputs, the same question is addressed for \textit{LK} and \textit{RLC}, and the prediction with the highest probability will be the model's prediction. The model uses the KGE to get all the triples probabilities after reification as mentioned earlier in \autoref{sec:methodology} and \autoref{fig:intention_prediction_pipeline}. During this instant, the model prediction is \textit{LK} as it has a higher probability than \textit{LLC} and \textit{RLC}.

Consider the TTC numerical value with the right following vehicle in the interval t=$[-6,-5]$, there is a drop of the TTC value from $-10$ to zero. This is due to the reason that initially, the TTC is calculated with respect to a right following vehicle instead of the one shown in \autoref{fig:frames_LLC_49_0638}a, but after some frames, the target vehicle completely passes the vehicle in the figure, and the TTC is now calculated with respect to that vehicle.

Moving on to the second instant which is described in \autoref{fig:frames_LLC_49_0638}b at $-4$ seconds before changing lanes. Despite that, the vehicle's lateral velocity and acceleration are zero. The model gives an LLC prediction (represented by a yellow triangle pointing to the left) because there is a medium-risk TTC with the preceding vehicle. Moreover, focusing on the interval t=$[-5,-4]$, it is shown that the right preceding vehicle TTC risk changed from low to medium, which makes the \textit{RLC} probability decrease, causing an increase in the \textit{LK} and \textit{LLC} probabilities.

After that, in the third captured frame represented in \autoref{fig:frames_LLC_49_0638}c. The target vehicle still has high-risk TTC with the preceding and right preceding vehicles. And the vehicle starts to accelerate in the left direction, moving with lateral velocity in the left direction as well. So, the vehicle started moving to merge and is about to change lanes. The model \textit{LLC} probability increased approximately to $90\%$.

Then, in the last capture after crossing the lane lines in \autoref{fig:frames_LLC_49_0638}d. The target vehicle is merging with right acceleration, left velocity, and high-risk right preceding vehicle. Moreover, by comparing the locations of the surrounding vehicles with respect to our target vehicle in\autoref{fig:frames_LLC_49_0638}c and \autoref{fig:frames_LLC_49_0638}d. It can be observed that the preceding vehicle with high-risk before changing lanes becomes the right preceding high-risk vehicle after the left lane changing. Also, the left preceding vehicle becomes the preceding vehicle. The same applies to all other target vehicle neighbors which causes a high change in the TTC values.

Finally, \autoref{tab:media} contains links for some multimedia videos that provide results of different scenes including the scene discussed in this section.

\begin{table}[ht]
\renewcommand{\arraystretch}{1.5}
\caption{Different Multimedia for Better Visualisation.}
\begin{center}
\begin{tabular}{|c|c|}
\hline
      Scenario          & Link\\
\hline
Left Lane Change   & \url{https://youtu.be/jPFj3YstBzs}\\
\hline
Left Lane Change   & \url{https://youtu.be/F7BSMsAyerI}\\
\hline
Lane Keeping & \url{https://youtu.be/zavuxrzb3KY}\\
\hline
Right Lane Change & \url{https://youtu.be/7xzeycfmRkc}\\
\hline
Right Lane Change & \url{https://youtu.be/wLfE8PfAUgU}\\
\hline
\end{tabular}
\label{tab:media}
\end{center}
\end{table}

\section{Conclusions and Future Work}\label{sec:concliusions}
In this work, the problem of vehicle lane change prediction is addressed using explainable contextual linguistic information that describes the target vehicle state and introduces risk awareness by considering the TTC risk with the surrounding vehicles. The proposed solution consists of three phases: linguistic input generation, knowledge graph embedding, and Bayesian inference and prediction. The first phase generates linguistic input categories from numerical values based on some threshold limits. The second phase takes the linguistic inputs and forms a knowledge graph reified triples in the form of a CSV file and uses the Ampligraph library for knowledge graph embedding. In the third, the implementation of a fully inductive reasoning system based on KGEs is carried out using Bayesian inference. This phase uses the embeddings from phase two to calculate the probabilities of different reified triples which are used for Bayesian reasoning to generate the final prediction. The proposed model is evaluated at different instants before the target vehicle changes lanes. The results showed that the model can predict lane-changing intention two seconds earlier with an f1-score of 97.95\%, and three seconds earlier with an f1-score of 93.60\%. This shows the reliability and robustness of the model to keep the f1-score higher than 90\% for three seconds before changing lanes. Moreover, the model surpassed other recent works that used the HighD dataset, especially at two seconds when the other models' scores started to decrease and our model kept its high score without decreasing.


%



\section*{Acknowledgment}
This research has been funded by the HEIDI project of the European Commission under Grant Agreement: 101069538.

\ifCLASSOPTIONcaptionsoff
  \newpage
\fi



%

\bibliographystyle{IEEEtran}
\bibliography{IEEEabrv,bibtex/bib/IEEEexample}

%








\end{document}